\documentclass[letterpaper, 10pt, conference]{ieeeconf}
\IEEEoverridecommandlockouts
\usepackage{cite}
\usepackage{amsmath,amssymb,amsfonts}
\usepackage{algpseudocode}
\usepackage{algorithm}		
\usepackage{graphicx}
\usepackage{textcomp}
\usepackage{xcolor}
\DeclareMathOperator*{\argmax}{argmax}
\DeclareMathOperator*{\argmin}{argmin}

\begin{document}
\onecolumn
\noindent Pre-print version of article Multi-Process Fusion: Visual Place Recognition Using Multiple Image Processing Methods, which has been published in Robotics and Automation Letters.

\vspace{\baselineskip}

\noindent Please cite this paper as:

\vspace{\baselineskip}

\noindent S. Hausler, A. Jacobson and M. Milford, ``Multi-Process Fusion: Visual Place Recognition Using Multiple Image Processing Methods," IEEE Robotics and Automation Letters, vol. 4, no. 2, pp. 1924-1931, April 2019. doi: 10.1109/LRA.2019.2898427

\vspace{\baselineskip}

\noindent bibtex:

\vspace{\baselineskip}

\noindent @ARTICLE\{Hausler2019,\\
  \indent author = \{S. \{Hausler\} and A. \{Jacobson\} and M. \{Milford\}\},\\
  \indent journal = \{IEEE Robotics and Automation Letters\},\\
  \indent title = \{Multi-Process Fusion: Visual Place Recognition Using Multiple Image Processing Methods\},\\
  \indent year = \{2018\},\\
  \indent volume = \{4\},\\
  \indent number = \{2\},\\
  \indent pages = \{1924-1931\},\\
  \indent doi = \{10.1109/LRA.2019.2898427\},\\
  \indent ISSN = {2377-3766\},\\
  \indent month = {April\},\\
\}

\twocolumn

\title{Multi-Process Fusion: Visual Place Recognition Using Multiple Image Processing Methods
\thanks{All authors are at the Queensland University of Technology. SH is supported by a Research Training Program Stipend and ARC Future Fellowship FT140101229. AJ is supported by an Advance Queensland Innovation Partnership, Caterpillar and Mining3. MM is with the Australian Centre for Robotic Vision and was partially supported by an ARC Future Fellowship FT140101229. Contact: stephen.hausler@hdr.qut.edu.au}
}


\author{Stephen Hausler, Adam Jacobson and Michael Milford}

\maketitle

\begin{abstract}
Typical attempts to improve the capability of visual place recognition techniques include the use of multi-sensor fusion and integration of information over time from image sequences. These approaches can improve performance but have disadvantages including the need for multiple physical sensors and calibration processes, both for multiple sensors and for tuning the image matching sequence length. In this paper we address these shortcomings with a novel ``multi-sensor" fusion approach applied to multiple image processing methods for a \emph{single} visual image stream, combined with a dynamic sequence matching length technique and an automatic processing method weighting scheme. In contrast to conventional single method approaches, our approach reduces the performance requirements of a single image processing methodology, instead requiring that within the suite of image processing methods, at least one performs well in any particular environment. In comparison to static sequence length techniques, the dynamic sequence matching technique enables reduced localization latencies through analysis of recognition quality metrics when re-entering familiar locations. We evaluate our approach on multiple challenging benchmark datasets, achieving superior performance to two state-of-the-art visual place recognition systems across environmental changes including winter to summer, afternoon to morning and night to day. Across the four benchmark datasets our proposed approach achieves an average F1 score of 0.96, compared to 0.78 for NetVLAD and 0.49 for SeqSLAM. We provide source code for the multi-fusion method and present analysis explaining how superior performance is achieved despite the multiple, disparate, image processing methods all being applied to a single source of imagery, rather than to multiple separate sensors.  

\smallskip
\end{abstract}


\section{Introduction}

\begin{figure}[h]
\centering
\includegraphics[width=\linewidth,keepaspectratio,trim=12cm 2.1cm 6.9cm 1.5cm,clip]{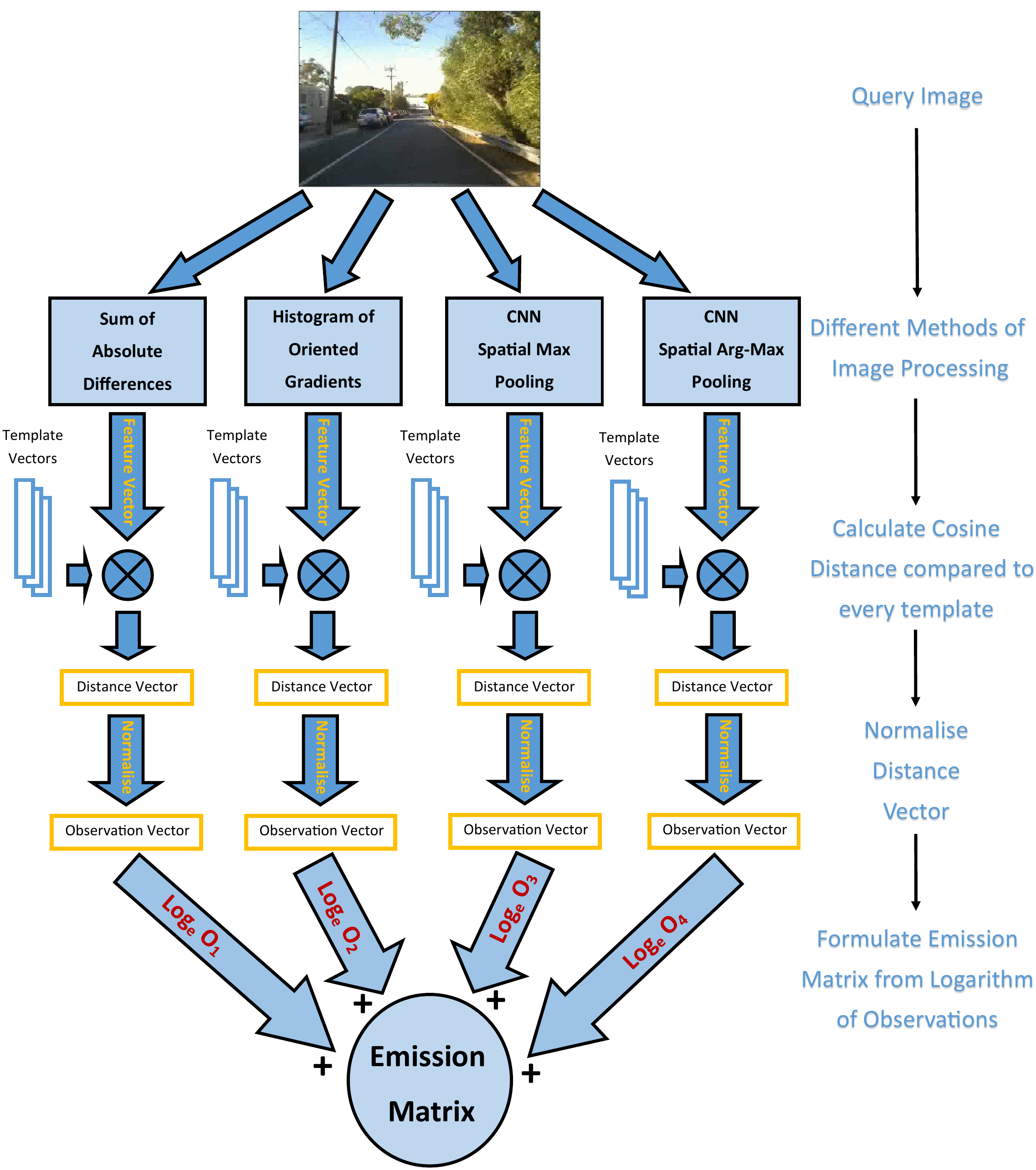}	
\caption{This diagram shows the formulation of the emission matrix using four methods of image processing. A feature vector is extracted for each method, which is then compared to the database templates to produce a distance vector. Logarithms and normalization are used to format each vector before creating the emission matrix.}
\label{EMatrix}
\end{figure}

Localization, a sub-component of Simultaneous Localization and Mapping, is challenging due to the unpredictability and diversity of real world environments. Both single and multi-sensor approaches have been used for localization, and in the subset of vision-only navigation, many individual image processing methods have been established. 

Convolutional Neural Networks (CNNs) have emerged as a means of learning a representation of an image and have excelled in object classification and scene recognition \cite{IM2015}. CNNs have successfully been used in place recognition as a replacement for traditional hand-crafted features \cite{SN2015}. In two recent approaches, HybridNet \cite{CZ2017} and NetVLAD \cite{AR2018}, a condition invariant representation is learnt to improve the place recognition performance. Sequence-based methods have also recently emerged: in Sequence SLAM \cite{MM2012}, patch-normalized, resolution-reduced images are compared using the sum of absolute differences (SAD) method over sequences of images. However, because the SAD method is susceptible to perceptual aliasing, long sequence lengths are required. To produce a more robust representation of the environment, more sophisticated hand-crafted features, like SURF \cite{BH2008} and HOG \cite{DN2005}, have been used for the task of visual localization \cite{CM2008,NT2018}.

In robotic navigation systems, multiple sensors are often deployed to overcome the limitations of a single sensor \cite{JA2015}. In this paper, we develop a unique approach for combining multiple image processing methods, akin to performing multi-sensor fusion, except with source imagery from just a single sensor. The rationale for such an approach is that different visual processing methods are better suited to different types of environments and conditions: combining these disparate processing methods can improve localization better than a ``one size fits all" to visual processing. To further improve the effectiveness of the system, we develop a new dynamic sequence length matching technique, utilizing a Hidden Markov Model (HMM) and novel variant to the Viterbi algorithm \cite{Viterbi} to determine the most likely location of the robot. Our approach dynamically adjusts the length of the sequence of recent images and sets the initial state estimate of the HMM, based on the location where the robot estimates that it has returned from an un-recognized scene back to a familiar scene.

Our approach imposes a novel set of requirements upon the image processing method: rather than reliance on an optimally performing single method, our multi-approach method simply requires that the multiple processing methods between them exhibit good performance across the range of deployed environments, reducing the performance requirements of any single processing method. In this paper, we demonstrate this functionality using four different image processing methods which have been demonstrated in the literature to have varying performance characteristics:
\begin{itemize}
\item{SAD with patch normalisation \cite{MM2012,Smart}.}
\item{Histogram of Oriented Gradients \cite{NT2018,HOGExample}.}
\item{CNN features, extracted from multiple spatial regions with a feature map \cite{CZ2017,Zetao2018}.}
\item{CNN features, only using the spatial coordinates of maximum activations \cite{Lost2018}.}
\end{itemize}
Using these methods, this paper presents four original contributions as follows:
\begin{itemize}
\item{A novel method for combining multiple methods of image processing across a sequence of images, using a Hidden Markov Model.}
\item{A method for computing a dynamic sequence matching length that calculates the rate of change in recognition quality within a set of recent images to find the optimal sequence sub-set to use in the HMM.}
\item{An alternative image comparison method that uses CNN maximum activation coordinates rather than activation magnitudes, and unlike \cite{Lost2018}, utilizing both the x and y coordinates of these maximally activated locations.}
\item{A method for automatically selecting the best image processing methods for the current environment, dubbed Multi-Process Fusion. The source code is available online\footnote{https://github.com/StephenHausler/Multi-Process-Fusion}.}
\end{itemize}

The paper proceeds as follows. In Section II, we review sequence-based localization methods and previous approaches in multi-sensor fusion. Section III presents our approach, describing the process we use to analyze images using multiple methods. Section IV details the setup of our experimental datasets and Section V evaluates their performance, with comparison to state-of-the-art methods. Section VI discusses these results and provides suggestions for future work.

\section{Related Work}

Visual place recognition has recently been a topic of intense research activity \cite{AR2018,ConvNet,Smart}. FAB-MAP \cite{CM2008} is a noteworthy example of an appearance-only localization approach, which uses a recursive Bayesian model to estimate location over a series of past observations, where the observations are the probability of observing a particular visual feature in the environment. SeqSLAM \cite{MM2012}, also a purely appearance-based method, demonstrates the discriminative power of temporal sequences of images, such that simple image matching techniques combined with image enhancement techniques can localize from day to night and summer to winter \cite{NP2013}. SeqSLAM fits a linear trajectory to the recent image difference scores, however this limitation has been addressed by using a Hidden Markov Model to find a non-linear trajectory through the image difference matrix \cite{Hansen}. Conditional Random Fields \cite{CRC} and using a network flow in a directed graph structure \cite{NT2018} have also been used to find the optimal route through the difference matrix. Unfortunately all sequence-based approaches suffer the limitation of localization latency proportional to the sequence length. Approaches that dynamically adjust the sequence length, such as leveraging GPS priors \cite{SeqGPS} or modelling the place hypotheses across a varying sequence length \cite{BJ2017}, are advantageous for localization in dynamic environments.

Utilizing multiple sources of information for navigation has been demonstrated by using multiple sensors, such as LIDAR, Sonar, RGB-Depth and Wi-Fi \cite{JA2018,MoreMultiSensor,MultiSensor3}. Fusing multiple sensors has been performed using probability models \cite{Zhang}, feature vector concatenation of normalized sensor data \cite{JA2013}, and multiplying normalized data across Gaussian-distributed clusters \cite{JA2018}. However, these multi-sensor approaches have the disadvantages of requiring expensive hardware and additional calibration requirements.

The concept of combining multiple methods of image processing, instead of using multiple sensors, has had limited investigation. In a prior work, multiple image processing methods were concatenated into a merged feature vector, with which a Convex Optimization Problem over sequences of images was used to determine the best match \cite{ZH2016}. An improved approach was later proposed, whereby the same convex problem was reframed to optimize the choice of modality rather than the choice of templates in a sequence \cite{SRAL}. Their results showed consistent performance over multiple datasets, since their solution utilizes the best processing methods for each dataset. However, their absolute performance was limited by the lack of sequential information. Rather than combining different image processing techniques, the authors of \cite{MultiLayerCNN} fuse multiple layers of a CNN for improved performance on an image retrieval task.

Rather than formulating feature vectors from the magnitude of CNN activations, prior works have also utilized the spatial position of activations within the feature map space. Yandex et al. adds a weighting onto sum pooling based on the position of each feature inside the feature map, with features closest to the center of the map space receiving the greatest weight \cite{Yandex15}. In another work, combining maximum activation coordinates with semantic information has been used to localize across opposite viewpoints \cite{Lost2018}. However, in \cite{Lost2018}, only the x-coordinate of the maximum activation keypoints are used to formulate the match score for a set of candidate locations. Since many autonomous navigation applications contain limited vertical viewpoint variations, it would be worthwhile to compare keypoints using both the x and y coordinates within each feature map.

\section{Proposed Approach}
Our proposed method uses a selection of proven image processing techniques and combines these together in a Hidden Markov Model (HMM) to determine the optimal estimated location over a sequence of images. We use a unique variant of the Viterbi algorithm to determine the estimated location over a dynamic sequence length and evaluate place matches using a novel variant of the trajectory scoring method used in SeqSLAM \cite{MM2012}.

\subsection{Image Processing Methods}

In the experiments in Section V, we use four different image processing methods: Sum of Absolute Differences (SAD), Histogram of Oriented Gradients (HOG) and two types of deep learnt features. As mentioned previously, the system presented in this paper is not rigidly tied to these specific four techniques, but rather to a suite of techniques with complementary performance characteristics in different environments. For SAD, we reduce the resolution of the images to 64$\times$32 pixels and perform patch normalization. For HOG \cite{DN2005}, we extract a vector of gradients out of a 640$\times$320 pixel input image with a cell size of 32$\times$32 pixels.  For the third method, we use the standard best practice method of extracting features out of a CNN, utilizing the innovations of two recent publications \cite{GS2018,CZ2017}. Our final method uses the coordinates of maximum activations within a convolutional layer. For all four methods, the cosine distance metric is used to compare these gradient vectors to the stored database template vectors.

\subsection{Deep Learnt Feature Matching}

In this sub-section, the two different methods of using deep learnt features will be explained in greater detail. Using HybridNet \cite{CZ2017}, we extract feature map activations from the Conv-5 ReLu layer. Pyramid pooling is used to reduce the feature vector dimensionality while maintaining the key features in each feature map. Each feature map is reduced to a five dimension long vector, containing the maximum activation across the whole feature map and the maximum activation in each quadrant of the feature map. Inspired by \cite{GS2018}, we use normalization to improve the discriminative power of these dimension-reduced activations. Each feature score is normalized by subtracting the mean and dividing by the standard deviation of that particular feature map across all past images.

For the second deep learnt method, again we use the Conv-5 layer, except the activation values themselves are discarded. Instead, scenes are compared using the (x,y) coordinate positions of the maximum activations in the current feature map and each template, using the Euclidean distance:
\begin{equation}
	D(k) = \frac{\sum^F_{f=1} \sqrt{(T_k(x_f) - I(x_f))^2 + (T_k(y_f) - I(y_f))^2}}{F}
\end{equation}
where D(k) is the distance metric from the current image I to template k. 

The pairwise distance between feature maps is summed for every feature map and normalized by the number of feature maps, to provide a score between the current image and each template. If two maximum activations are in the same spatial location within a particular feature map, then this feature map will have a high similarity score. Inherently this method will have greater luminosity and seasonal invariance, but inferior viewpoint invariance. This offset is reduced by combining this processing method with more viewpoint invariant processing methods, such as traditional CNN features. After computing multiple image difference vectors, normalization is required to merge these vectors together, as described in the subsequent section.

\subsection{Configuration of the Observation and Transition Matrices}
Each observation matrix is a normalized image difference matrix. Each new image difference vector is initially normalized between -0.001 and 0.999, where 0.999 is the best matching template. Normalized values under $O_{thresh}$ are then floored to 0.001 – this is an additional penalty to discourage matches when one of the observations significantly disagrees with the other observations:
\begin{equation}
O(k,i) = \left\{
	\begin{aligned}
		& \frac{\text{max}(D) - D(k)}{\text{max}(D) - \text{min}(D)} - \epsilon		&&	\text{if } O(k,i) \geq O_{thresh}\,,
	\\	& \epsilon																&&	\text{else}\,.
	\end{aligned}
\right.
\end{equation}
\[
	\epsilon = 0.001 
\]
where $D$ is the difference vector between the current image and every database template, $i$ is the current image number and $k$ is the template number. The minimum (0.001) and maximum (0.999) values were chosen to be values near 0 and 1 for which the natural logarithm can be applied without causing the resultant value to asymptote to 0 or $\infty$. The number of decimal places was limited so that in cases of extreme perceptual aliasing, the discrimination between these places becomes purely on the difference between image processing methods rather than miniscule differences in the cosine distance scores.

The transition matrix is defined by the assumption that the autonomous vehicle, during a query traverse through an environment, will maintain a velocity between zero and five times the velocity during the reference traverse. Therefore we set $V_{min}$ to 0 frames and $V_{max}$ to 5 frames.
\begin{equation}
T(k,i) = \left\{
	\begin{aligned}
		& \log (1)		&& \text{if } V_{min} \leq (k-i) \leq V_{max}\,,
	\\	& \log (\epsilon)	&& \text{else}\,.
	\end{aligned}
\right.
\end{equation}

This equation has the effect of discouraging discontinuous jumps through the matrix of pseudo-probability scores, by increasing the scores for templates outside the velocity window.

\subsection{Hidden Markov Model across Sequences of Images}
The HMM is applied across a sequence of recent images. The emission matrix is formulated by combining up to $M$ observations, with the assumption that all observations are conditionally independent. In reality there is a conditional dependence based on the underlying visual information that is common between the different representations of the same scene, however we still achieve state-of-the-art results while making this assumption. We use the logarithm of the observation values and then perform addition during the Viterbi algorithm. Refer to Fig. \ref{EMatrix} for a flow chart of the process from query image to emission matrix. The emission matrix, $E$, is defined by: 
\begin{equation}
E = \sum_{n=1}^{M} \log O_n\label{eq}      
\end{equation}
where each observation $O_1$ to $O_M$ is a normalized similarity score between the most recent images in the query traverse and every image in the reference traverse, for each image processing method. For the transition matrix $T$ and emission sequence over time $E$ = (e\textsubscript{1},e\textsubscript{2},..,e\textsubscript{$\tau$}), the optimal hidden state sequence $S$ is determined using the Viterbi algorithm \cite{Viterbi}. In our approach, we treat the Viterbi algorithm as a cost function, rather than as a probability function. The specific state probability is not required for localization, only the positions of the minimum costs and the ratio between costs.

Our method of computing the Viterbi algorithm is described in pseudocode below:
\vskip\medskipamount 
\leaders\vrule width \linewidth\vskip1pt 
\vskip\medskipamount 
\nointerlineskip
\begin{algorithmic}[1] 
	\State Initialise $D$ to the first emission
	\For{$i = 2$ to $\tau$}
		\State $D[k,i] = \underset{1 \leq k \leq N}{\max}( D[k,i-1] + T) + E$
		\State $H[k,i] = \underset{1 \leq k \leq N}{\argmax}(D[k,i-1] + T)$
	\EndFor
	\State $S = \underset{1 \leq k \leq N}{\argmax}(D[k,\tau])$
	\For{$k = \tau,\tau - 1, .. , 1$}
		\State $S(k-1) = H(S(k),k)$
	\EndFor
\end{algorithmic}
\vskip\medskipamount 
\leaders\vrule width \linewidth\vskip1pt 
\vskip\medskipamount 
\nointerlineskip
\noindent where N is the number of database templates, $\tau$ is the sequence length and S is the optimal hidden state sequence.

Unlike a standard Viterbi algorithm, our approach removes the initial probability term. Using the pseudo-probability from the previous pass through the algorithm encourages a match near the prior estimated location, however this method increases the re-localization delay after losing localization and increases the risk of the system entering a sequence of false positives. To offset the removal of the initial probability, we utilize a dynamic sequence start location and a dynamic sequence length. When the robot returns from an unfamiliar to a familiar scene, re-localization does not occur until a number of frames equal to the sequence length has elapsed. By dynamically reducing the sequence length, the localization delay disadvantage of SeqSLAM is reduced. Our approach performs a preliminary search within a larger, fixed-length, sequence to find a change in the recognition quality, so that the sequence used in our HMM consists of an improved sub-set of the larger sequence. The dynamic sequence begins at the largest negative rate of change $Q_{ROC}$ in a quality score $Q$ across a maximum sequence length $S_{max}$.

\begin{equation}
\begin{aligned}
	Q_{ROC} = \text{min}(\delta Q_k)			&&	k\in [S_{min},S_{max}]			\label{eq}
\end{aligned}
\end{equation}
\begin{equation}
\text{seqStart} = \left\{
	\begin{aligned}
		& \argmin(\delta Q_k)	&&	\text{if } \left\lvert Q_{ROC} \right\rvert \geq Q_t\,,	  
	\\    & 0		      			&& \text{else}\,.
	\end{aligned}
\right.
\end{equation}

For our experiments, we set $S_{min}$ to 5 frames and $S_{max}$ to 20 frames. $Q_t$ is a threshold that ensures the gradient is large enough to indicate a variation in the scene characteristics. This approach causes the HMM sequence to start after the return from an unfamiliar or aliased scene (a large quality score denotes a poor match) to a familiar or distinctive scene (a smaller quality score). By measuring the rate-of-change in the quality score, we can detect where these transitions occur within a sequence.

The quality score, inspired by SeqSLAM \cite{MM2012}, is a ratio between the maximum score in a column of the emission matrix and the next maximum score outside a window around the first maximum. Because the emission matrix contains values that are the logarithm of the original normalized observations, this quality ratio becomes non-linear with respect to the original scores. To explain the motivation for this innovation, an example will be employed. If best match has a normalized score of 0.999, and the next minimum score is 0.9, then there is some perceptual aliasing. If, at the next time step, the next minimum score is 0.99, then significant perceptual aliasing is present. In SeqSLAM, the difference in ratio between 0.999 and 0.9 and 0.999 and 0.99 is minimal, however by non-linearizing the quality scoring using logarithms, a significant difference is added between these ratio values. With this, the dynamic sequence length is able to detect severe perceptual aliasing in large datasets.

After the Viterbi algorithm computes the optimal path through the pseudo-probability matrix, the quality score is recalculated, by computing the ratio between the value at the optimal path and the next largest value outside a window around the optimal path. These scores are summed for the entire sequence and then normalized by sequence length, to generate an averaged quality score. This quality score is used to differentiate a familiar scene from a novel scene.

\subsection{Multi-Process Fusion Algorithm (MPF)}

To provide a single place recognition algorithm that can be deployed across a wide range of environments, the choice of processing methods must be dynamically selected to suit the conditions. To this end we have developed a final sub-system, which compares the matching performance of each processing method individually and removes the worst performer, using a voting method. We take the indices of the largest normalised score for the current frame for each processing method and find the similar template index between them. The processing method with a single-frame location hypothesis that is furthest from this averaged template index is assumed to be the worst performing method and is filtered out. When the Viterbi algorithm is applied in Multi-Process Fusion, the emission matrix is re-calculated at each index of the sequence, only using the best processing methods for that particular image.

\section{Experimental Method}

We demonstrate our approach on four established benchmark datasets, which have been extensively tested in recent literature \cite{SRAL,NT2018,GS2018,JA2018}. Each dataset is briefly described in the sub-paragraphs below and displayed in Fig. \ref{MatchCompare}. 

\textbf{St Lucia} – consists of multiple car traverses through the suburb of St Lucia, Brisbane across five different times of day \cite{LuciaDatasetRef}. We use the early morning traverse as the reference dataset and the late afternoon video as the query, with significant appearance change occurring between morning and afternoon. We use the first 4000 frames of the original 15 FPS video, which is more than a complete loop around the suburb of St Lucia. The dataset provides GPS ground truth and we use a ground-truth tolerance of 30 meters for the results. 

\textbf{Nordland} – The Nordland dataset \cite{NordlandDatasetRef} is recorded from a train travelling for 728 km through Norway across four different seasons. We use the Summer and Winter traverses and extract the first one hour and 40 minutes out of the video at 1 FPS. Sections of the dataset where the train is stopped or in a tunnel are excluded, resulting in a frame count of 4150 frames. In this dataset, each frame in one traverse is in the same location as the corresponding frame in another traverse. As a result, we perform ground-truth checks by comparing the query traverse frame number to the matching database frame number, with a ground-truth tolerance of 10 frames. 

\textbf{Oxford RobotCar} - RobotCar was recorded over a year across different times of day, seasons and routes \cite{RobotCar}. We use the first 2.5km of a route through Oxford, matching from an overcast day (2014-12-09-13-21-02) to night on the next day (2014-12-10-18-10-50). This corresponds to 2050 frames because we down sample the original frame rate by a factor of three and start both traverses at the same location. We use a ground truth tolerance of 40 meters, consistent with a recent publication \cite{GS2018}. 

\textbf{University Campus} – to evaluate the system on a non-vehicular dataset, the university campus dataset was chosen. This dataset was created by recording a video using a web-camera on a laptop while walking around the QUT campus \cite{JA2018}. We use every third frames out of the original dataset, resulting in a query traverse size of 3226 frames. Both traverses are during the day, thus the appearance variation is low, however significant camera shake and moderate viewpoint variations are present. We use a ground truth tolerance of 10 meters, as per a similar study \cite{JA2018}.

\section{Results}

This section presents the results, evaluating the performance trends across a suite of image processing front ends. Refer to Table \ref{Experiments} for a summary of the processing combinations that we evaluate. Precision-Recall curves are used to compare the combinations to each other and to the benchmarks OpenSeqSLAM2 \cite{SeqSLAM2} and NetVLAD \cite{AR2018}. SeqSLAM was setup using a sequence length of 20 frames and a linear trajectory search velocity from 0.8 to 1.2. For NetVLAD, we used the CNN pre-trained on Pittsburgh 30k. We apply NetVLAD into our HMM model, so that NetVLAD is advantaged by using sequences of images. In Fig. \ref{F1Bar}, our best algorithm, Multi-Process Fusion (MPF), is compared to SeqSLAM and NetVLAD using the F1 score metric. On three of the four datasets, our approach achieves the highest F1 score and we achieve almost identical performance to NetVLAD on the Campus dataset. 

\begin{table}[h]
\caption{List of Image Processing Combinations}
\label{Experiments}
\centering
\begin{tabular}{|c|l|}
\hline\hline
\bfseries Name & \bfseries Description\\
\hline
1 Observation - CNN & Spatial max pooling CNN features at Conv-5\\
\hline
1 Observation - CNN-D & Argmax CNN features at Conv-5\\
\hline
1 Observation - HOG & HOG at cell size 32$\times$32 pixels\\
\hline
1 Observation - SAD & One observation using patch-normalised SAD\\
\hline
2 Obs - CNN, CNN-D & Spatial max and argmax CNN features\\
\hline
2 Obs - CNN, HOG &	Spatial max CNN combined with HOG\\
\hline
2 Obs - HOG, SAD &	Combining the two hand-crafted methods\\
\hline
4 Obs - SAD, HOG, &	All four image processing methods \\
CNN, CNN-D & \\
\hline
MPF	  & Multi-Process Fusion applied to four methods\\
\hline
4 Obs - CNN	 & Multiple CNN layers from Conv-3 to 6\\
\hline
\end{tabular}
\end{table}

\begin{figure}[h]
\centering
\includegraphics[width=\linewidth,trim=3.2cm 1.2cm 3.2cm 1.0cm,clip]{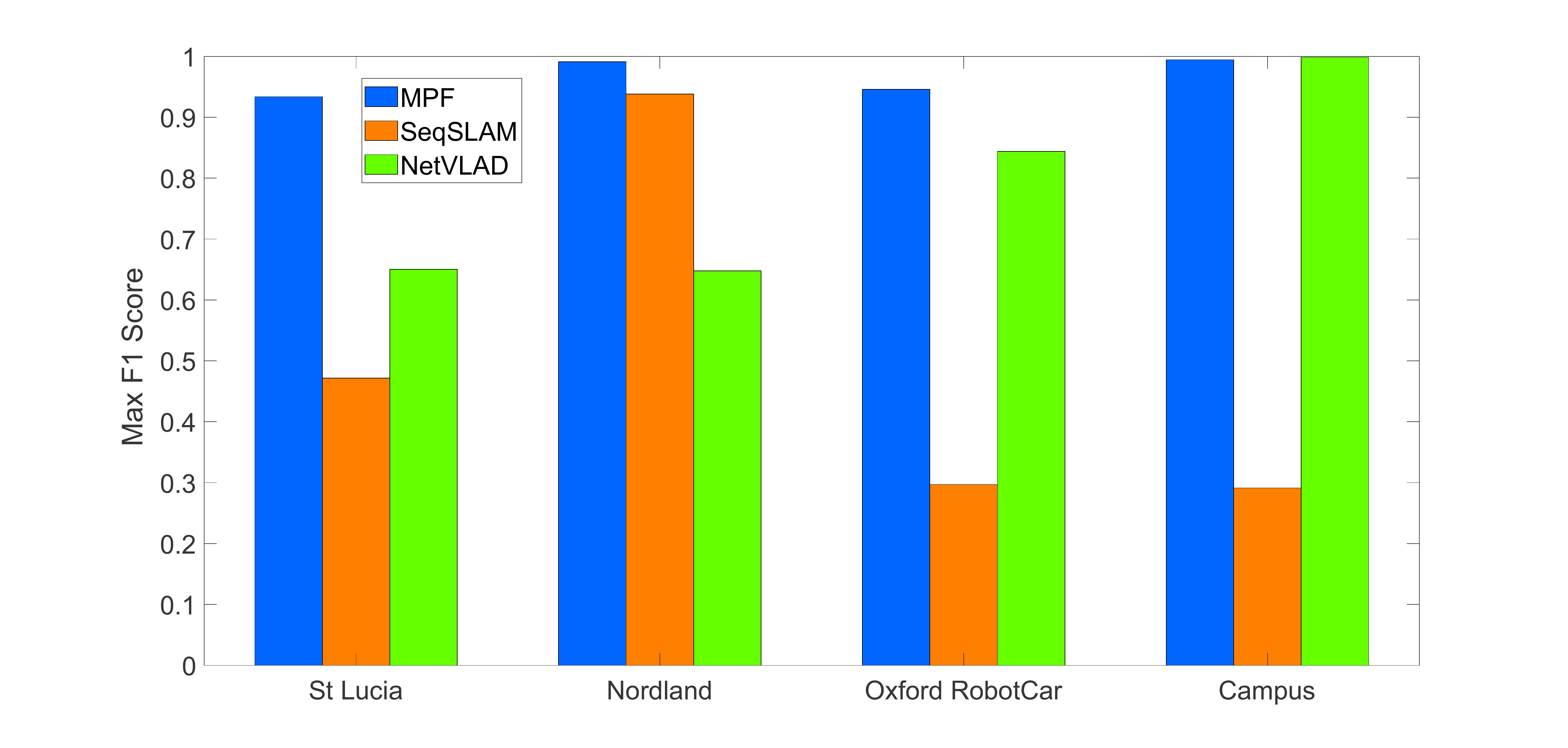}	
\caption{Maximum F1 score for Multi-Process Fusion (MPF), SeqSLAM and NetVLAD on St Lucia, Nordland, Oxford RobotCar and Campus.}
\label{F1Bar}
\end{figure}

\begin{figure}[h]
\centering
\includegraphics[width=\linewidth,trim=3.2cm 0.5cm 3.2cm 0cm,clip]{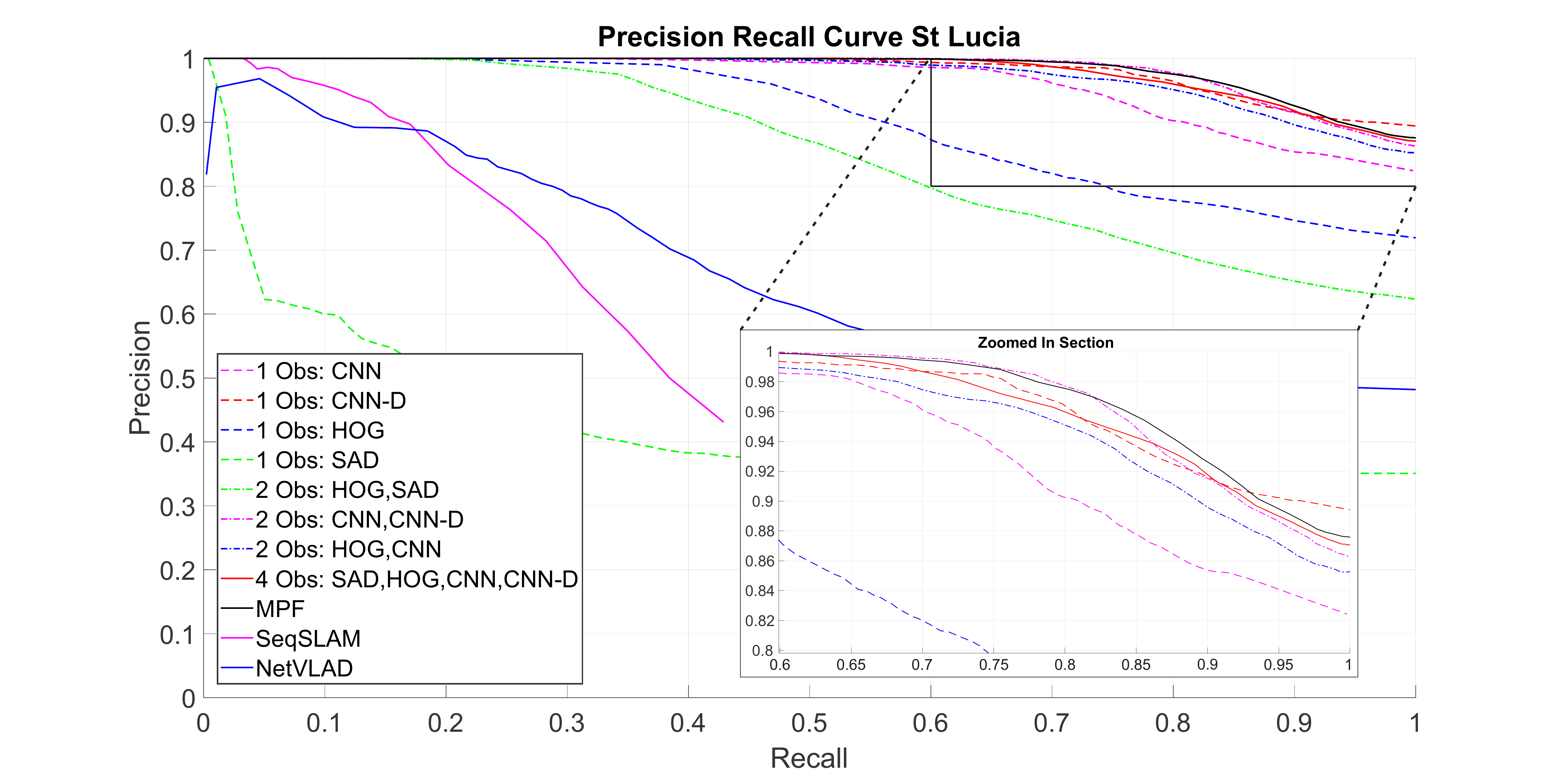}	
\caption{Precision-Recall curves for multiple combinations of processing methods on the St Lucia dataset. Dashed is a single method, dot-dashed is two methods and a solid line denotes four methods. Best viewed in color.}
\label{PRLucia}
\end{figure}

\subsection{St Lucia}

The results in Fig. \ref{PRLucia} illustrates the precision-recall performance improving as the number of observations increases, with Multi-Process Fusion achieving 47\% recall at 100\% precision. We achieve a higher precision-recall result than both SeqSLAM and NetVLAD. SeqSLAM is limited by its reliance on the Sum of Absolute differences method, which is shown to be the worst performing image comparison method on this dataset. NetVLAD, which was trained on Pittsburgh, a highly urban environment, has difficulty localizing in a low-density suburb with severe illumination variations. This dataset has minimal viewpoint variation, thus using the spatial coordinates of maximum activations in each feature map (termed CNN-D in Fig. \ref{PRLucia}) performs better than any other single image processing method.

\begin{figure}[h]
\centering
\includegraphics[width=\linewidth,trim=3.2cm 0.5cm 3.2cm 0.8cm,clip]{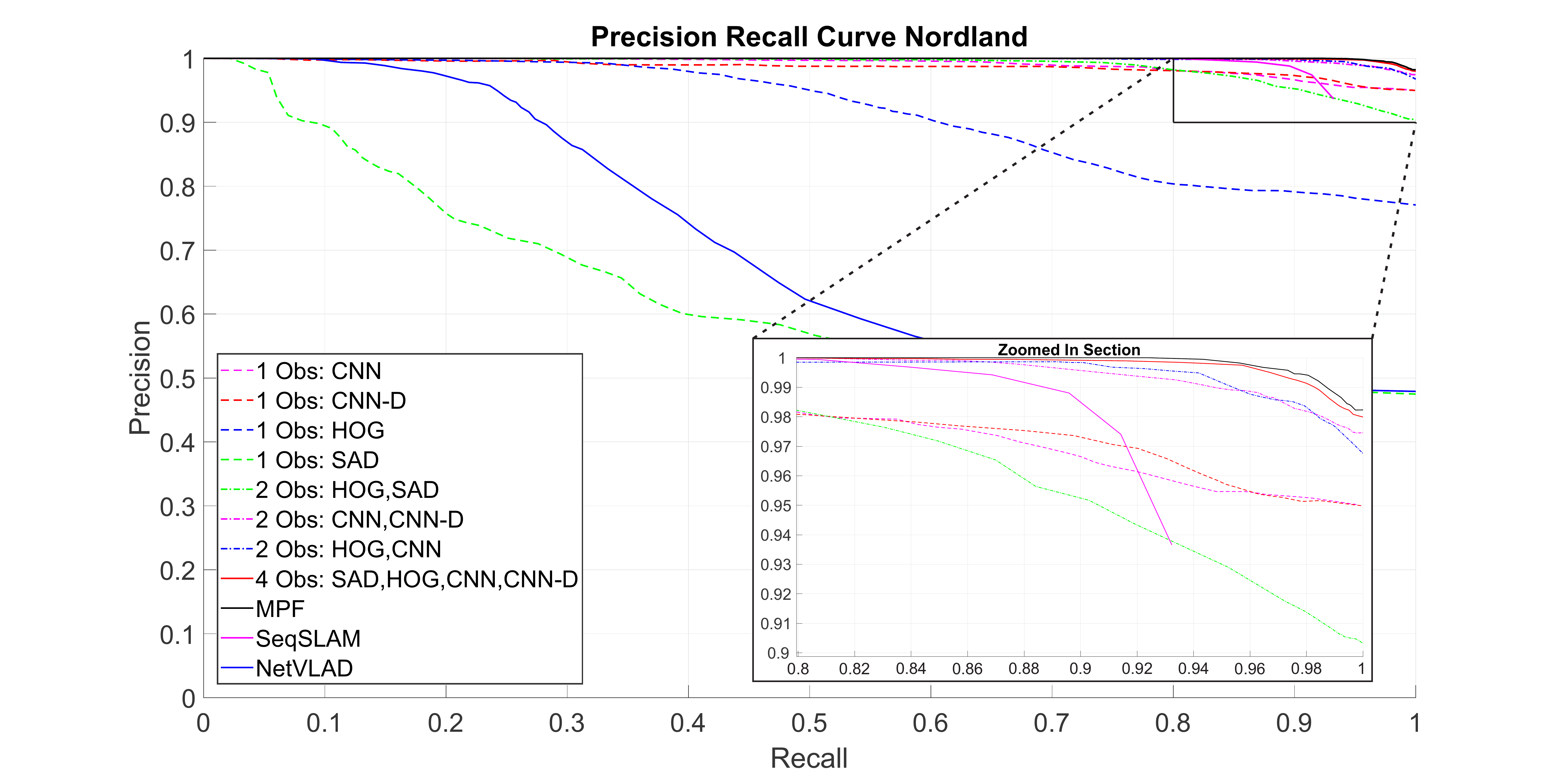}	
\caption{Precision-Recall curves for multiple combinations of processing methods on the Nordland dataset. Dashed is a single method, dot-dashed is two methods and a solid line denotes four methods. Best viewed in color.}
\label{PRNord}
\end{figure}

\subsection{Nordland}

As observed in Fig. \ref{PRNord}, increasing the number of observations significantly improves the localization performance. On this dataset Multi-Process Fusion achieves 92\% recall at 100\% precision and SeqSLAM achieves 70\% recall at 100\% precision. This dataset is unique in that there are no viewpoint variations and the velocity is constant for large sections of the dataset, which causes sequence-based approaches to excel. It is inferred that the performance of NetVLAD suffers due to the lack of urban scenery on the Nordland dataset, since NetVLAD was trained on Pittsburgh, an urban environment. 

\begin{figure}[h]
\centering
\includegraphics[width=\linewidth,trim=3.2cm 0.5cm 3.2cm 0cm,clip]{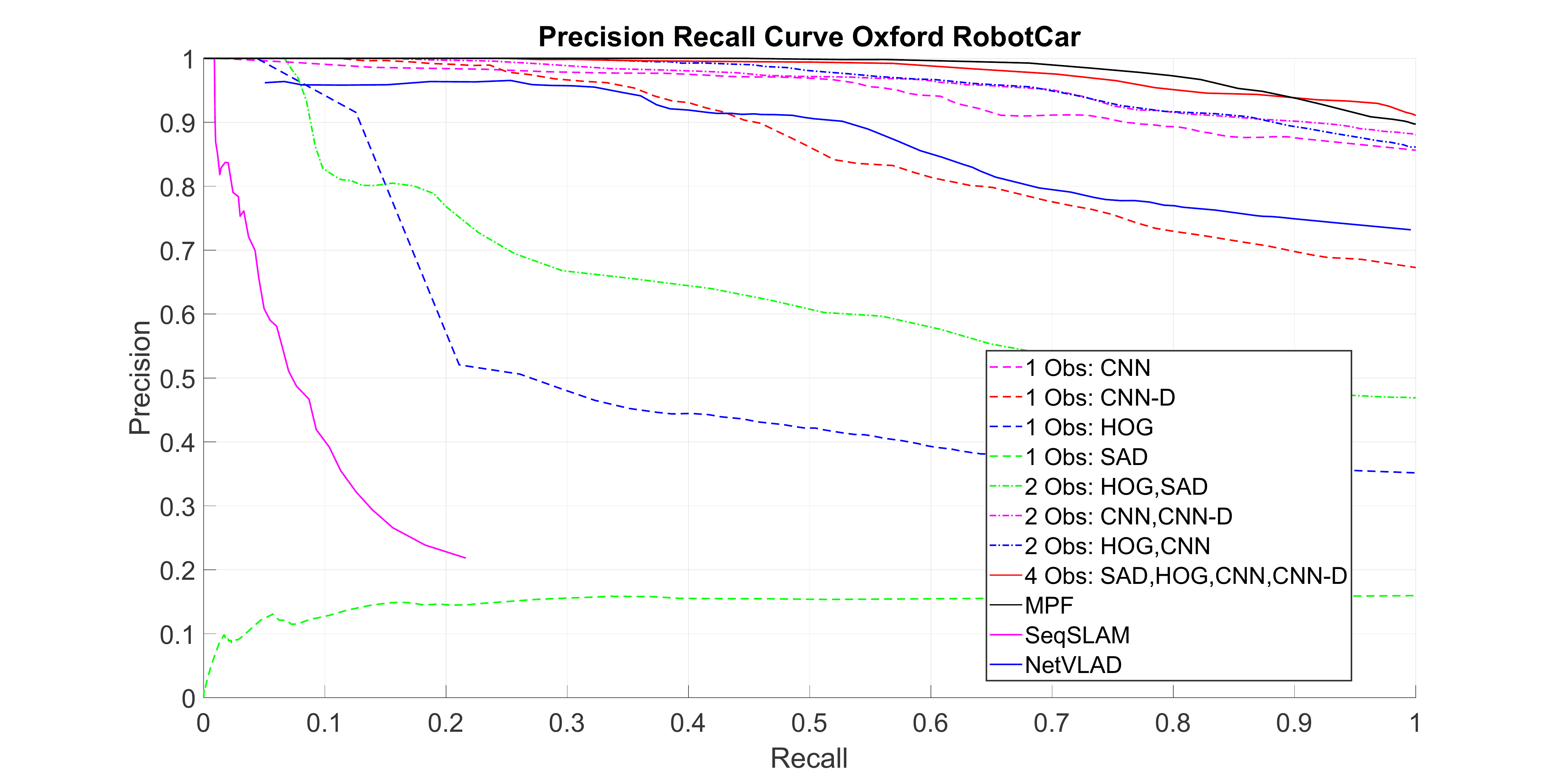}	
\caption{Precision-Recall curves for multiple combinations of processing methods for Oxford RobotCar. Dashed is a single method, dot-dashed is two methods and a solid line denotes four methods. Best viewed in color.}
\label{PROxford}
\end{figure}

\subsection{Oxford RobotCar}

In Fig. \ref{PROxford}, again using multiple image processing methods achieves the best results. This is a challenging dataset, matching from day to night, thus hand-crafted features perform poorly. CNNs trained on images over time are able to recognize features between day and night, resulting in the best Precision-Recall curve from NetVLAD so far. Multi-Process Fusion is a further improvement and achieves 44\% recall at 100\% precision. We hypothesize that while standalone hand-crafted features fail, when hand-crafted features are combined with deep learnt features, the two approaches complement each other and offset each other's weaknesses. On this dataset we include all stops - we process every frame irrespective of the vehicle's velocity. This is a further challenge that causes SeqSLAM's trajectory search method to fail.

\begin{figure}[h]
\centering
\includegraphics[width=\linewidth,trim=3.2cm 0.5cm 3.2cm 0.8cm,clip]{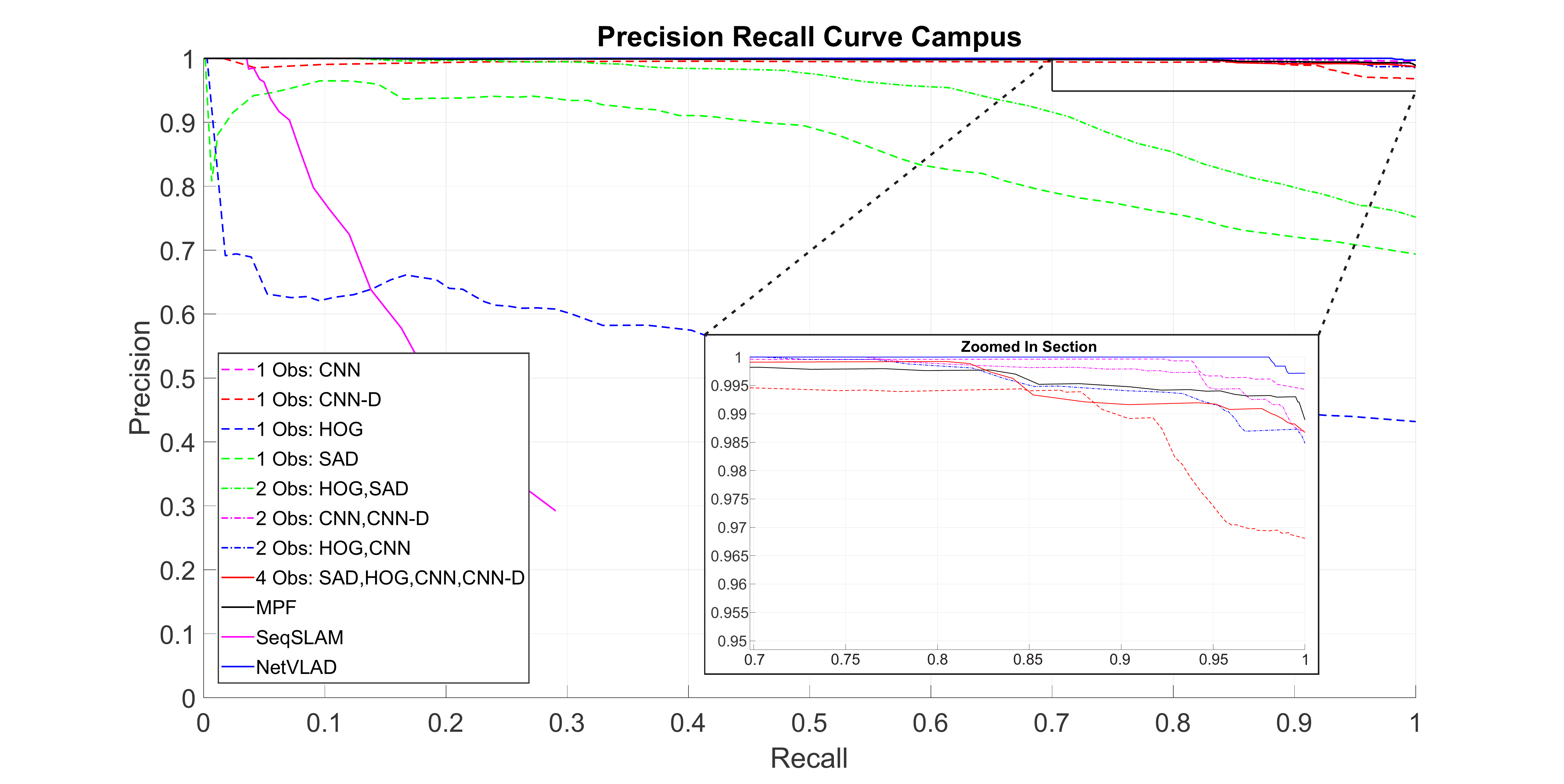}	
\caption{Precision-Recall curves for multiple combinations of processing methods on the Campus dataset. Dashed is a single method, dot-dashed is two methods and a solid line denotes four methods. Best viewed in color.}
\label{PRCampus}
\end{figure}

\subsection{Campus}

This campus dataset has little condition variation, but moderate viewpoint variations due to camera shake. This impacts the performance of global descriptors such as SAD and HOG, but here viewpoint-invariant CNN approaches yield almost perfect recognition, as shown in Fig. \ref{PRCampus}. 1 Obs: CNN achieves over 99\% precision at 100\% recall and out-performs Multi-Process Fusion. Because the viewpoint variations affect both HOG and SAD, and Multi-Process Fusion only filters a single image processing method, Multi-Process Fusion cannot achieve the performance of the CNN methods. The high viewpoint variations impacts CNN-D, since the spatial position of the maximum activations are changing with respect to viewpoint, however CNN-D still outperforms the global descriptors of SAD and HOG. SeqSLAM performs poorly on this dataset, with 3.5\% recall at 100\% precision, due to the variations in viewpoint. On this dataset, sequence NetVLAD is marginally superior to our approach, and achieves 98\% precision at 100\% recall. Our recall at 100\% precision is reduced by a false positive that occurs on a perceptually aliased stairwell in this dataset (see Fig. \ref{MatchCompare}). We discovered that this false positive could be removed by using multiple layers of a CNN, which detect additional features in the images and enables differentiation between the two stairwells.

\subsection{Investigation into Combining Multiple Layers of a CNN}

Because of the exceptional performance of CNNs on the Campus dataset, we conducted an experiment to apply our approach to concatenate multiple layers of a CNN. We use Conv-3 through to Conv-6 of HybridNet and each layer is treated as a separate image processing method with their own sets of reference database templates. In Fig. \ref{PRCCCC}, we show the Precision-Recall curves using this method for the four datasets. On the Nordland and RobotCar datasets, combining hand-crafted features and deep learnt features is superior in terms of recall at 100\% precision; however on the Campus dataset, we achieve a recall of 98\% at 100\% precision, matching the results achieved by NetVLAD. This experiment demonstrates our approach on a different set of visual features, by utilizing multiple layers of a CNN. 

\begin{figure}[h]
\centering
\includegraphics[width=\linewidth,trim=3.2cm 0.5cm 3.2cm 0cm,clip]{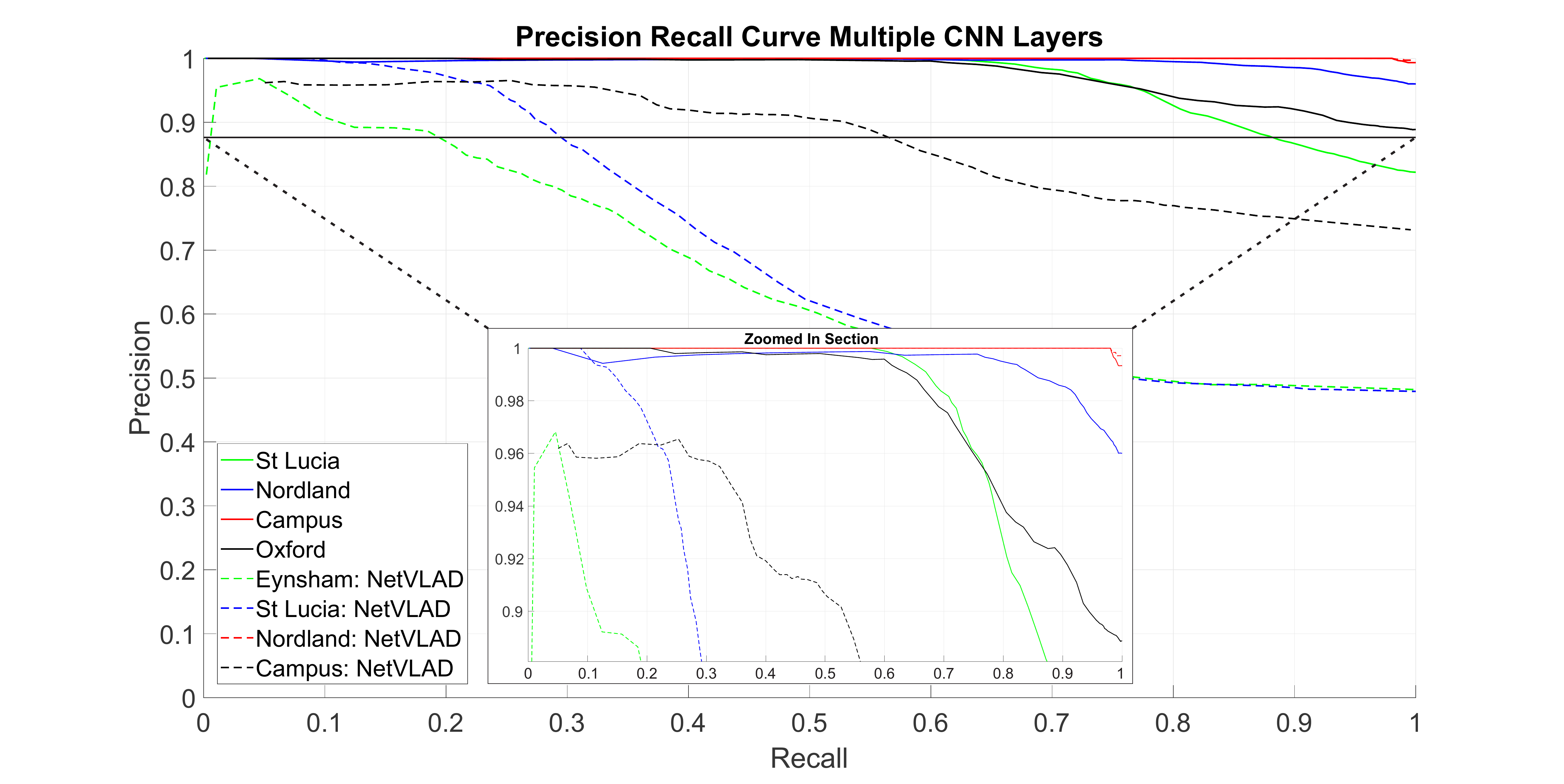}	
\caption{Combining multiple CNN layers using a HMM on all five datasets, compared to NetVLAD. Solid line denotes our method, dashed line is NetVLAD. Best viewed in color.}
\label{PRCCCC}
\end{figure}

\section{Discussion and Conclusion}

This paper proposes a unique approach to combining multiple methods of image processing for visual place recognition, akin to performing multi-sensor fusion using just a single sensor. This is achieved using a Hidden Markov Model to find the optimal location estimate from multiple observations over a dynamic sequence of images. Our approach enables reliable place recognition across a variety of difficult conditions, with key results including 92\% recall at 100\% precision on the Nordland dataset from Summer to Winter and 44\% recall at 100\% precision for day to night matching on the Oxford RobotCar dataset. Using a dynamic sequence length reduces the localization delay disadvantage, with a minimum sequence length of just 5 frames using our approach. Because the dynamic sequence length is computed based on a preliminary analysis of the recognition quality, if the robot travels from a unique environment back to a previously traversed environment, the sequence length will typically reduce (depending on the severity of the perceptual aliasing) and the system will quickly re-localize. We also contribute an alternative method of extracting features out of a CNN, by discarding the activation values and simply using the maximum activation coordinates in each feature map. Normally such a method would be prohibitive due to viewpoint variations; however we claim that this method can be used in our approach because we couple this method with a more viewpoint invariant method, like traditional CNN features. 

On all four datasets, the addition of multiple processing methods typically improves the precision-recall performance. Certain combinations of processing methods perform better; combining deep learnt features with hand-crafted features is typically superior to only using hand-crafted features. Adding more observations is not necessarily an improvement, as with the Campus dataset, where a single CNN observation achieves the best performance. It is the combination of the optimal processing methods for the specific environmental conditions that is required for practical autonomy. Our algorithm Multi-Process Fusion achieves this and we provide an example of the voting weights in Fig. \ref{ProcessRank}. In Fig. \ref{MatchCompare}, we show several examples where Multi-Process Fusion successfully localizes and both SeqSLAM and NetVLAD generate an incorrect match. 

\begin{figure}[h]
\centering
\includegraphics[width=0.95\linewidth,keepaspectratio,trim=1.0cm 0.0cm 0.5cm 0.0cm,clip]{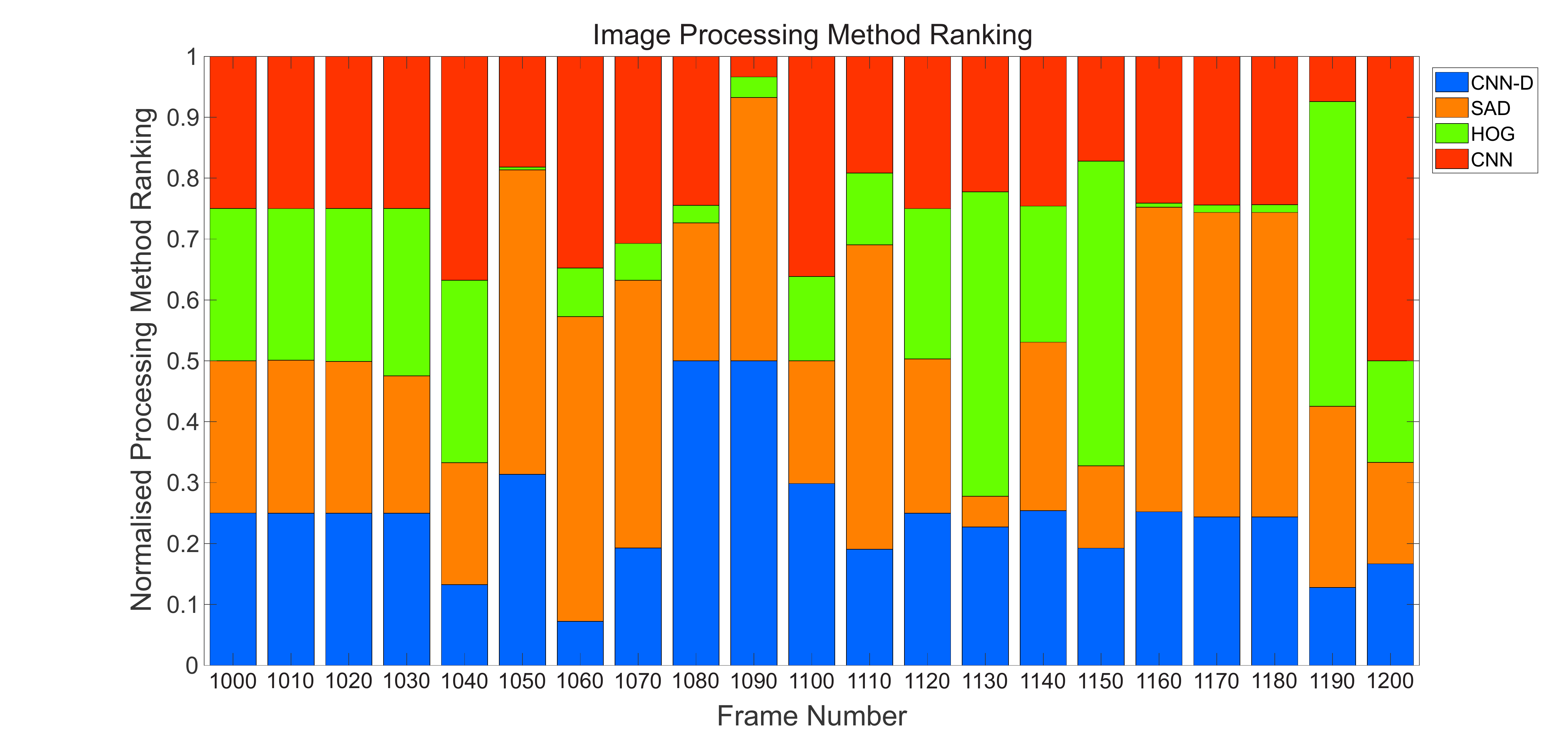}	
\caption{Frame-by-frame image processing ranking using Multi-Process Fusion, from a subset of the St Lucia dataset. This stacked bar graph shows the proportional ranking between methods for 21 selected frames. A larger segment within a column indicates the worst processing method for that particular frame.}
\label{ProcessRank}
\end{figure}

Improving the computational efficiency would further improve our current work. Rather than using multiple processing methods simultaneously, instead multiple methods would only be used when the system detects a reduction in the matched scene quality score, iteratively adding additional processing methods until either the match confidence improves or a novel environment is identified. Our results demonstrate a considerable advantage when compared to localization with a single image processing method, which enables future autonomous applications to gain the advantages of a multi-sensor approach without requiring multiple sensors, potentially reducing the financial cost of autonomous navigation.

\begin{figure}[h]
\centering
\includegraphics[width=0.8\linewidth,keepaspectratio,trim=0.0cm 3.6cm 0.0cm 1.3cm,clip]{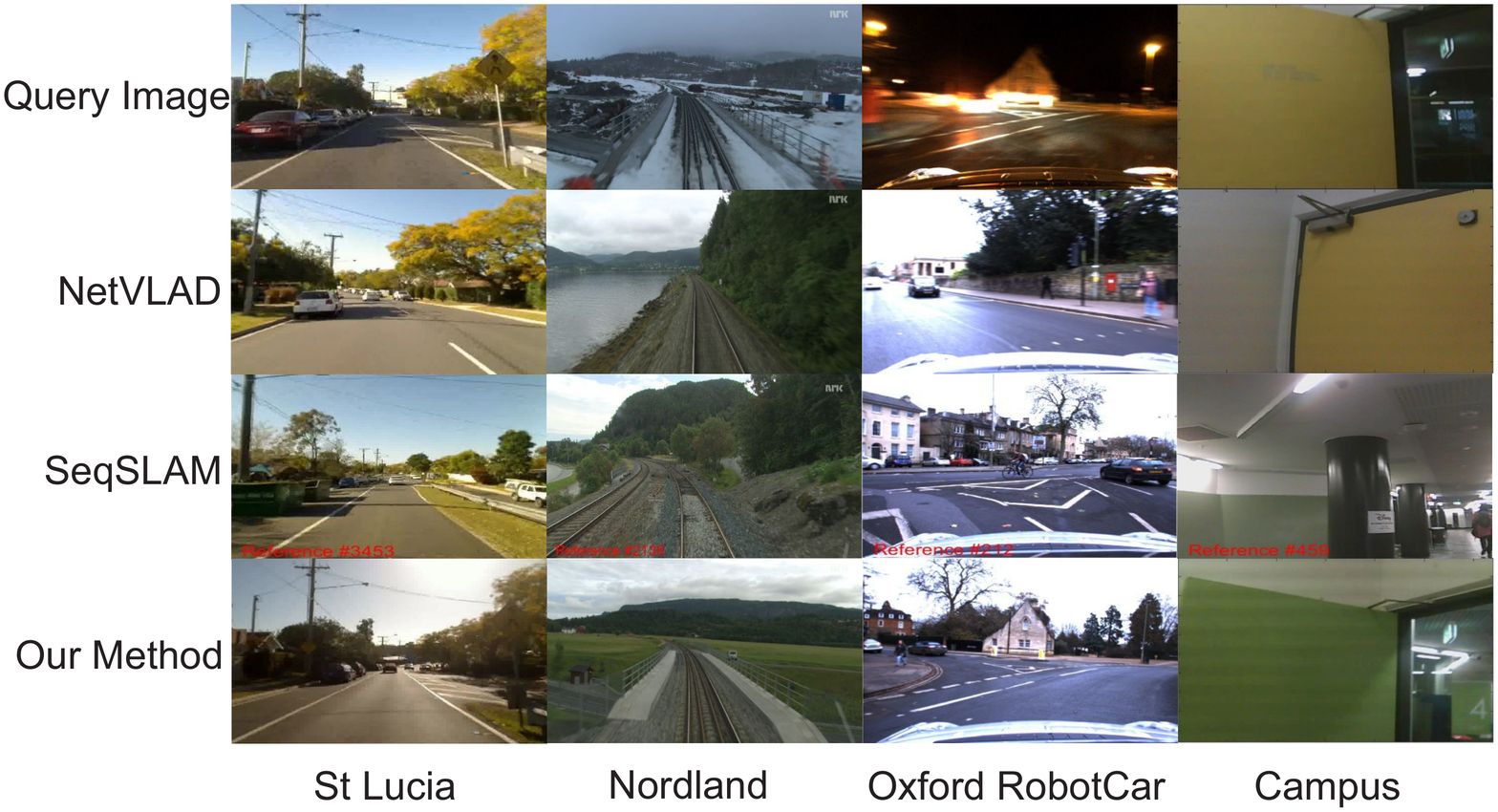}	
\caption{The first three columns show examples where our method successfully localizes and both NetVLAD and SeqSLAM produce incorrect matches. The fourth column displays a failure case for Multi-Process Fusion on the Campus dataset.}
\label{MatchCompare}
\end{figure}

\bibliographystyle{IEEEtran}
\bibliography{MultiProcessFusion}

\begin{thebibliography}{10}
\providecommand{\url}[1]{#1}
\csname url@samestyle\endcsname
\providecommand{\newblock}{\relax}
\providecommand{\bibinfo}[2]{#2}
\providecommand{\BIBentrySTDinterwordspacing}{\spaceskip=0pt\relax}
\providecommand{\BIBentryALTinterwordstretchfactor}{4}
\providecommand{\BIBentryALTinterwordspacing}{\spaceskip=\fontdimen2\font plus
\BIBentryALTinterwordstretchfactor\fontdimen3\font minus
  \fontdimen4\font\relax}
\providecommand{\BIBforeignlanguage}[2]{{%
\expandafter\ifx\csname l@#1\endcsname\relax
\typeout{** WARNING: IEEEtran.bst: No hyphenation pattern has been}%
\typeout{** loaded for the language `#1'. Using the pattern for}%
\typeout{** the default language instead.}%
\else
\language=\csname l@#1\endcsname
\fi
#2}}
\providecommand{\BIBdecl}{\relax}
\BIBdecl

\bibitem{IM2015}
O.~Russakovsky, J.~Deng, H.~Su, J.~Krause, S.~Satheesh, S.~Ma, Z.~Huang,
  A.~Karpathy, A.~Khosla, M.~Bernstein, A.~C. Berg, and L.~Fei-Fei, ``Imagenet
  large scale visual recognition challenge,'' \emph{International Journal of
  Computer Vision}, vol. 115, no.~3, pp. 211--252, 2015.

\bibitem{SN2015}
N.~Sünderhauf, S.~Shirazi, F.~Dayoub, B.~Upcroft, and M.~Milford, ``On the
  performance of convnet features for place recognition,'' in \emph{2015
  IEEE/RSJ International Conference on Intelligent Robots and Systems
  (IROS)}.\hskip 1em plus 0.5em minus 0.4em\relax IEEE, 2015, pp. 4297--4304.

\bibitem{CZ2017}
Z.~Chen, A.~Jacobson, N.~Sünderhauf, B.~Upcroft, L.~Liu, C.~Shen, I.~Reid, and
  M.~Milford, ``Deep learning features at scale for visual place recognition,''
  in \emph{2017 IEEE International Conference on Robotics and Automation
  (ICRA)}.\hskip 1em plus 0.5em minus 0.4em\relax IEEE, 2017, pp. 3223--3230.

\bibitem{AR2018}
R.~Arandjelović, P.~Gronat, A.~Torii, T.~Pajdla, and J.~Sivic, ``Netvlad: Cnn
  architecture for weakly supervised place recognition,'' \emph{IEEE
  Transactions on Pattern Analysis and Machine Intelligence}, vol.~40, no.~6,
  pp. 1437--1451, 2018.

\bibitem{MM2012}
M.~J. Milford and G.~F. Wyeth, ``Seqslam: Visual route-based navigation for
  sunny summer days and stormy winter nights,'' in \emph{2012 IEEE
  International Conference on Robotics and Automation}.\hskip 1em plus 0.5em
  minus 0.4em\relax IEEE, 2012, pp. 1643--1649.

\bibitem{BH2008}
H.~Bay, A.~Ess, T.~Tuytelaars, and L.~Van~Gool, ``Speeded-up robust features
  (surf),'' \emph{Computer Vision and Image Understanding}, vol. 110, no.~3,
  pp. 346--359, 2008.

\bibitem{DN2005}
N.~Dalal and B.~Triggs, ``Histograms of oriented gradients for human
  detection,'' in \emph{2005 IEEE Computer Society Conference on Computer
  Vision and Pattern Recognition (CVPR'05)}, vol.~1, 2005, pp. 886--893 vol. 1.

\bibitem{CM2008}
M.~Cummins and P.~Newman, ``Fab-map: Probabilistic localization and mapping in
  the space of appearance,'' \emph{The International Journal of Robotics
  Research}, vol.~27, no.~6, pp. 647--665, 2008.

\bibitem{NT2018}
T.~Naseer, W.~Burgard, and C.~Stachniss, ``Robust visual localization across
  seasons,'' \emph{Robotics, IEEE Transactions on}, vol.~34, no.~2, pp.
  289--302, 2018.

\bibitem{JA2015}
A.~Jacobson, Z.~Chen, and M.~Milford, ``Autonomous multisensor calibration and
  closed-loop fusion for slam,'' \emph{Journal of Field Robotics}, vol.~32,
  no.~1, pp. 85--122, 2015.

\bibitem{Viterbi}
A.~Viterbi, ``Error bounds for convolutional codes and an asymptotically
  optimum decoding algorithm,'' \emph{IEEE Transactions on Information Theory},
  vol.~13, no.~2, pp. 260--269, 1967.

\bibitem{Smart}
E.~Pepperell, P.~I. Corke, and M.~J. Milford, ``All-environment visual place
  recognition with smart,'' in \emph{2014 IEEE International Conference on
  Robotics and Automation (ICRA)}.\hskip 1em plus 0.5em minus 0.4em\relax IEEE,
  2014, pp. 1612--1618.

\bibitem{HOGExample}
C.~McManus, B.~Upcroft, and P.~Newmann, ``Scene signatures : localised and
  point-less features for localisation,'' in \emph{Proceedings of Robotics
  Science and Systems (RSS)}, Conference Proceedings.

\bibitem{Zetao2018}
Z.~Chen, L.~Liu, I.~Sa, Z.~Ge, and M.~Chli, ``Learning context flexible
  attention model for long-term visual place recognition,'' \emph{IEEE Robotics
  and Automation Letters}, vol.~3, no.~4, pp. 4015--4022, 2018.

\bibitem{Lost2018}
S.~Garg, N.~Sünderhauf, and M.~Milford, ``Lost? appearance-invariant place
  recognition for opposite viewpoints using visual semantics,''
  \emph{Proceedings of Robotics: Science and Systems XIV}, 2018.

\bibitem{ConvNet}
N.~Sünderhauf, S.~Shirazi, F.~Dayoub, B.~Upcroft, and M.~Milford, ``On the
  performance of convnet features for place recognition,'' in \emph{2015
  IEEE/RSJ International Conference on Intelligent Robots and Systems
  (IROS)}.\hskip 1em plus 0.5em minus 0.4em\relax IEEE, 2015, pp. 4297--4304.

\bibitem{NP2013}
P.~Neubert, N.~Sünderhauf, and P.~Protzel, ``Appearance change prediction for
  long-term navigation across seasons,'' in \emph{2013 European Conference on
  Mobile Robots}, 2013, pp. 198--203.

\bibitem{Hansen}
P.~Hansen and B.~Browning, ``Visual place recognition using hmm sequence
  matching,'' in \emph{2014 IEEE/RSJ International Conference on Intelligent
  Robots and Systems}.\hskip 1em plus 0.5em minus 0.4em\relax IEEE, 2014, pp.
  4549--4555.

\bibitem{CRC}
C.~Cadena, D.~Galvez-López, J.~D. Tardos, and J.~Neira, ``Robust place
  recognition with stereo sequences,'' \emph{IEEE Transactions on Robotics},
  vol.~28, no.~4, pp. 871--885, 2012.

\bibitem{SeqGPS}
O.~Vysotska, T.~Naseer, L.~Spinello, W.~Burgard, and C.~Stachniss, ``Efficient
  and effective matching of image sequences under substantial appearance
  changes exploiting gps priors,'' in \emph{2015 IEEE International Conference
  on Robotics and Automation (ICRA)}.\hskip 1em plus 0.5em minus 0.4em\relax
  IEEE, 2015, pp. 2774--2779.

\bibitem{BJ2017}
J.~Bruce, A.~Jacobson, and M.~Milford, ``Look no further: Adapting the
  localization sensory window to the temporal characteristics of the
  environment,'' \emph{IEEE Robotics and Automation Letters}, vol.~2, no.~4,
  pp. 2209--2216, 2017.

\bibitem{JA2018}
A.~Jacobson, Z.~Chen, and M.~Milford, ``Leveraging variable sensor spatial
  acuity with a homogeneous, multi-scale place recognition framework,''
  \emph{Biological Cybernetics}, pp. 1--17, 2018.

\bibitem{MoreMultiSensor}
I.~Jebari, S.~Bazeille, E.~Battesti, H.~Tekaya, M.~Klein, A.~Tapus, D.~Filliat,
  C.~Meyer, S.~Ieng, R.~Benosman, E.~Cizeron, J.~Mamanna, and B.~Pothier,
  ``Multi-sensor semantic mapping and exploration of indoor environments,'' in
  \emph{2011 IEEE Conference on Technologies for Practical Robot
  Applications}.\hskip 1em plus 0.5em minus 0.4em\relax IEEE, 2011, pp.
  151--156.

\bibitem{MultiSensor3}
J.~Collier, S.~Se, and V.~Kotamraju, ``Multi-sensor appearance-based place
  recognition,'' in \emph{2013 International Conference on Computer and Robot
  Vision}.\hskip 1em plus 0.5em minus 0.4em\relax IEEE, 2013, pp. 128--135.

\bibitem{Zhang}
P.~Zhang, J.~Gu, E.~E. Milios, and P.~Huynh, ``Navigation with imu/gps/digital
  compass with unscented kalman filter,'' in \emph{IEEE International
  Conference Mechatronics and Automation, 2005}, vol.~3.\hskip 1em plus 0.5em
  minus 0.4em\relax IEEE, pp. 1497--1502 Vol. 3.

\bibitem{JA2013}
M.~J. Milford and A.~Jacobson, ``Brain-inspired sensor fusion for navigating
  robots,'' in \emph{2013 IEEE International Conference on Robotics and
  Automation}.\hskip 1em plus 0.5em minus 0.4em\relax IEEE, 2013, pp.
  2906--2913.

\bibitem{ZH2016}
H.~Zhang, F.~Han, and H.~Wang, ``Robust multimodal sequence-based loop closure
  detection via structured sparsity,'' \emph{Proceedings of Robotics: Science
  and Systems XII}, vol.~12, 2016.

\bibitem{SRAL}
F.~Han, X.~Yang, Y.~Deng, M.~Rentschler, D.~Yang, and H.~Zhang, ``Sral: Shared
  representative appearance learning for long-term visual place recognition,''
  \emph{IEEE Robotics and Automation Letters}, vol.~2, no.~2, pp. 1172--1179,
  2017.

\bibitem{MultiLayerCNN}
Y.~Li, X.~Kong, H.~Fu, and Q.~Tian, ``Aggregating hierarchical binary
  activations for image retrieval,'' \emph{Neurocomputing}, 2018.

\bibitem{Yandex15}
A.~B. Yandex and V.~Lempitsky, ``Aggregating local deep features for image
  retrieval,'' in \emph{2015 IEEE International Conference on Computer Vision
  (ICCV)}.\hskip 1em plus 0.5em minus 0.4em\relax IEEE, 2015, pp. 1269--1277.

\bibitem{GS2018}
S.~Garg, N.~Sünderhauf, and M.~Milford, ``Don't look back: Robustifying place
  categorization for viewpoint- and condition-invariant place recognition,'' in
  \emph{2018 IEEE International Conference on Robotics and Automation (ICRA)},
  2018.

\bibitem{LuciaDatasetRef}
A.~J. Glover, W.~P. Maddern, M.~J. Milford, and G.~F. Wyeth, ``Fab-map +
  ratslam: Appearance-based slam for multiple times of day,'' in \emph{2010
  IEEE International Conference on Robotics and Automation}.\hskip 1em plus
  0.5em minus 0.4em\relax IEEE, 2010, pp. 3507--3512.

\bibitem{NordlandDatasetRef}
N.~Sünderhauf, P.~Neubert, and P.~Protzel, ``Are we there yet? challenging
  seqslam on a 3000km journey across all four seasons,'' in \emph{Proc. of
  Workshop on Long-Term Autonomy IEEE International Conference on Robotics and
  Automation (2013)}.\hskip 1em plus 0.5em minus 0.4em\relax IEEE, 2013.

\bibitem{RobotCar}
W.~Maddern, G.~Pascoe, C.~Linegar, and P.~Newman, ``1 year, 1000 km: The oxford
  robotcar dataset,'' \emph{The International journal of robotics research.},
  vol.~36, no.~1, pp. 3--15, 2017.

\bibitem{SeqSLAM2}
B.~Talbot, S.~Garg, and M.~Milford, ``Openseqslam2.0: An open source toolbox
  for visual place recognition under changing conditions,'' \emph{2018
  International Conference on Intelligent Robots and Systems (IROS)}, 2018.

\end{thebibliography}

\end{document}